\documentclass[runningheads]{llncs}

\usepackage{eccv}

\usepackage{eccvabbrv}

\usepackage{graphicx}
\usepackage{booktabs}

\usepackage[accsupp]{axessibility}  % Improves PDF readability for those with disabilities.

\usepackage[pagebackref,breaklinks,colorlinks,citecolor=eccvblue]{hyperref}
\usepackage{orcidlink}
\usepackage{hyperref}       % hyperlinks
\usepackage{url}            % simple URL typesetting
\usepackage{amsfonts}       % blackboard math symbols
\usepackage{nicefrac}       % compact symbols for 1/2, etc.
\usepackage{microtype}      % microtypography
\usepackage{enumitem}
\usepackage{amsmath}
\usepackage{multirow}
\usepackage{listings}
\usepackage{algorithm}
\usepackage{algorithmic}
\usepackage{marvosym}

\begin{document}

\title{Boosting the Power of Small Multimodal Reasoning Models to Match Larger Models with Self-Consistency Training} 

\titlerunning{Abbreviated paper title}

\author{Cheng Tan\inst{1,4,5\star} \and
Jingxuan Wei\inst{2,3\star}${^{\textrm{\Letter}}}$ \and Zhangyang Gao\inst{1,4,5\star} \and Linzhuang Sun\inst{2,3} \and Siyuan Li\inst{1,4,5} \and Ruifeng Guo\inst{2,3} \and Bihui Yu\inst{2,3} \and
Stan Z. Li\inst{4}${^{\textrm{\Letter}}}$\orcidlink{0000-0002-2961-8096}}

\authorrunning{Tan. et al.}

\institute{Zhejiang University \and
Shenyang Institute of Computing Technology, Chinese Academy of Sciences \and University of Chinese Academy of Sciences \and AI Lab, Research Center for Industries of the Future, Westlake University \and Institute of Advanced Technology, Westlake Institute for Advanced Study
\email{tancheng@westlake.edu.cn,weijingxuan20@mails.ucas.edu.cn}\\
$\star$ Equal contribution, \textrm{\Letter} Corresponding author
}

\maketitle

\begin{abstract}
Multimodal reasoning is a challenging task that requires models to reason across multiple modalities to answer questions. Existing approaches have made progress by incorporating language and visual modalities into a two-stage reasoning framework, separating rationale generation from answer inference. However, these approaches often fall short due to the inadequate quality of the generated rationales. In this work, we delve into the importance of rationales in model reasoning. We observe that when rationales are completely accurate, the model's accuracy significantly improves, highlighting the need for high-quality rationale generation. Motivated by this, we propose MC-CoT, a self-consistency training strategy that generates multiple rationales and answers, subsequently selecting the most accurate through a voting process. This approach not only enhances the quality of generated rationales but also leads to more accurate and robust answers. Through extensive experiments, we demonstrate that our approach significantly improves model performance across various benchmarks. Remarkably, we show that even smaller base models, when equipped with our proposed approach, can achieve results comparable to those of larger models, illustrating the potential of our approach in harnessing the power of rationales for improved multimodal reasoning. The code is available at \href{https://github.com/chengtan9907/mc-cot}{github.com/chengtan9907/mc-cot}.
\end{abstract}

\section{Introduction}
% chain-of-thought has been thoroughly investigated -> CoT is proved to be useful in multimodal reasoning -> the key to CoT is the quality of rationales, existing approaches are not good enough -> we propose a simple yet effective strategy that improves the quality of rationales -> our approach is able to boost the performance of smaller models to match larger models (moreover, prove that multimodal chain-of-thought can be benefited even using small models)
Recent advances in large language models~\cite{t5,opt,flan-t5,gpt-3,gpt-4,anthropic,instructgpt,llama,llama2,palm,palm2} have led to the exploration of Chain-of-Thought (CoT) prompting~\cite{cot,cot-sc,tot,got}. This approach, which directs the model to systematically unravel rationales before providing answers, rather than responding directly, has showcased the model's impressive efficacy across a variety of natural language processing (NLP) tasks. Moreover, the advent of CoT prompting has catalyzed a plethora of research endeavors delving into the reasoning prowess of large language models. A diverse range of Chain-of-Thought strategies has been investigated, including the voting-facilitated CoT-SC~\cite{cot-sc}, the transition from chain-like to tree-like thinking with Tree-of-Thoughts~\cite{tot}, and further expansion into graph-structured thinking~\cite{got}.

\begin{figure}[t]
\centering
\includegraphics[width=1.0\textwidth]{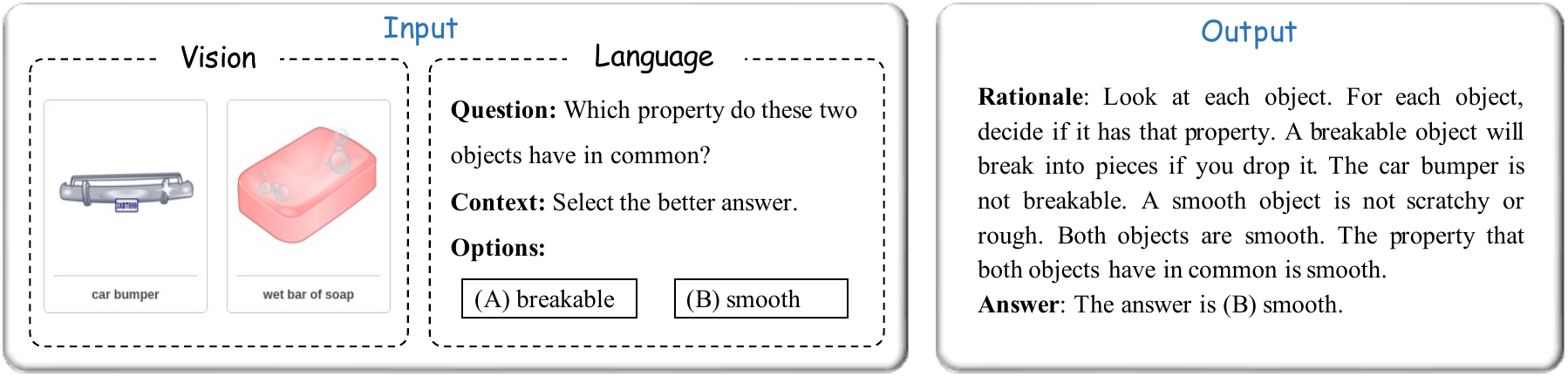}
\caption{An example of multimodal reasoning that answers the question by reasoning across both vision and language modalities.}
% \vspace{5mm}
\label{fig:example}
\end{figure}

While CoT reasoning has been thoroughly established in the realm of language models, its foray into the vast and intricate landscape of multimodal reasoning is still in its infancy. As shown in Fig.~\ref{fig:example}, Multimodal reasoning~\cite{scienceqa,visual-chatgpt,huang2022towards,llava,llama-adapter,lu2023chameleon,llama-adapterv2,liu2023mmbench,bai2023qwen,fu2023mme,yin2023survey,lavin}, which inherently involves the seamless fusion of information from disparate modalities such as text and images, presents unique challenges. The process of extracting, correlating, and generating rationales across multiple modalities is decidedly more complex than the tasks encountered in a solely text-based modality. A recent seminal work, Multimodal-CoT~\cite{mm-cot}, has pioneered the application of the Chain-of-Thought prompting strategy to multimodal reasoning tasks. This approach encompasses a two-stage framework that distinctively separates rationale generation from answer inference. By obliging the model to generate rationales prior to answering questions, Multimodal-CoT mirrors the language-only CoT prompting, thus enabling reasoning across multiple modalities.

\begin{figure}[!h]
  \vspace{-2mm}
  \centering
  \includegraphics[width=0.7\textwidth]{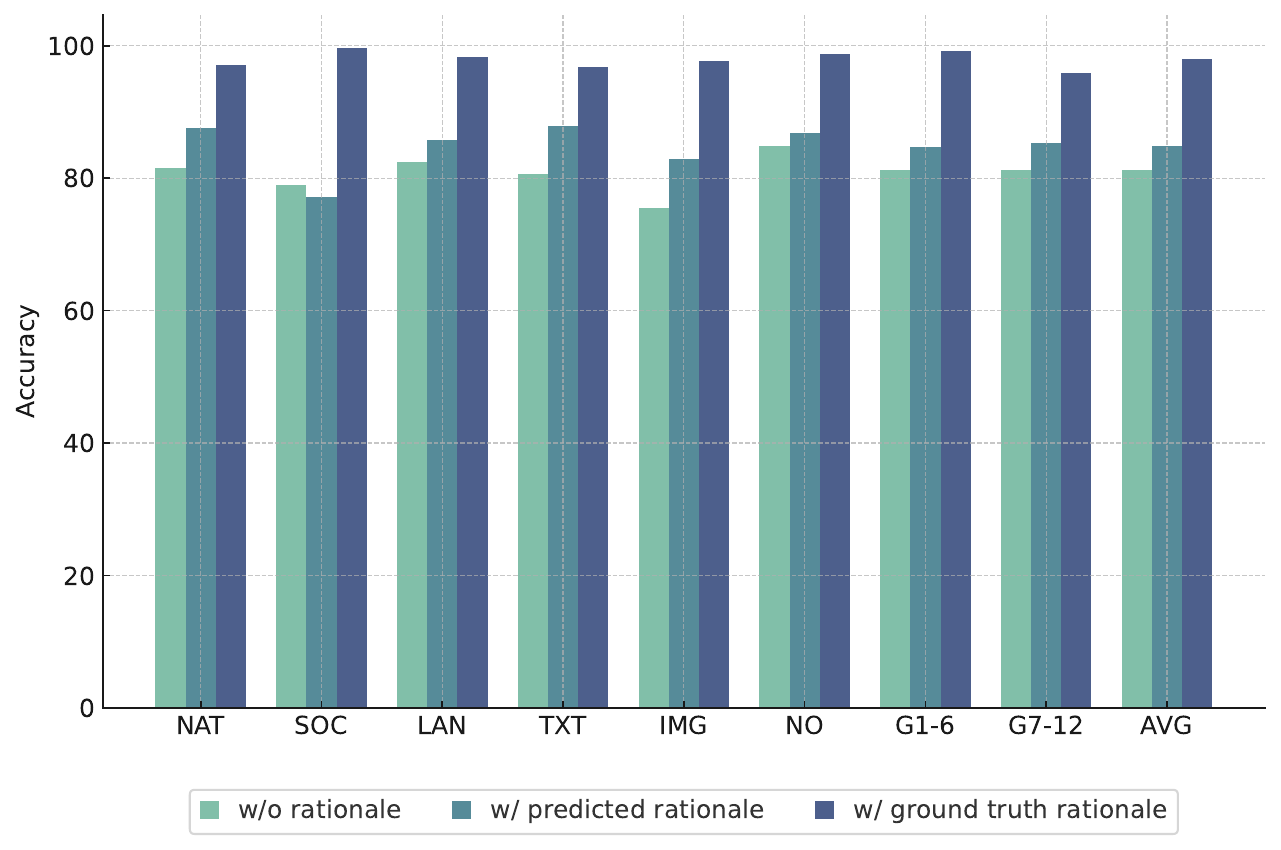}
  \caption{The comparison of answer accuracy on ScienceQA using the Multimodal-CoT framework with no rationale, predicted rationales, and ground-truth rationales.}
  \label{fig:rationale_comparison}
  \end{figure}

Despite Multimodal-CoT has made promising strides in the realm of multimodal reasoning, as evidenced in Fig.~\ref{fig:rationale_comparison}, its improvements over the no-rationale baseline are still limited. Moreover, compared to the ground-truth rationale, the predicted rationale falls short, often yielding results that lack relevance to the posed question. This discrepancy primarily stems from the quality of the generated rationales, highlighting the crucial role of rationale quality in the success of the chain-of-thought reasoning process.

In light of the above observations, it becomes evident that the potency of the Chain-of-Thought reasoning in a multimodal scenario is intrinsically tethered to the accuracy of the rationales generated. A high-quality rationale not only sheds light on the model's thought process but also paves the way for more precise answers. This leads to a critical question: \textit{how can we develop a strategy to enhance the quality of these rationales, and consequently, improve the overall performance of multimodal CoT?}

Our study is driven by the hypothesis that enhancing the quality of rationale generation can significantly improve the model's reasoning capabilities and overall performance. To this end, we introduce a simple yet effective strategy that capitalizes on the inherent variability of deep neural models during training, particularly stemming from mechanisms like dropout. Our approach involves having the model generate multiple rationales and then voting for the most consistent words across these rationales to yield a more refined and accurate rationale. The same voting mechanism is then applied to answer generation, further boosting the model's confidence in its predictions. It is important to note that the inference phase remains completely unaffected by this voting mechanism and continues to operate in the same manner as it does in the original Multimodal CoT framework. Through this approach, we aim to facilitate a more robust and accurate multimodal reasoning ability.

Extensive experiments across ScienceQA~\cite{scienceqa} and A-OKVQA~\cite{a-okvqa} benchmark datasets demonstrate the efficacy of our proposed approach. Notably, by improving the rationale quality, even smaller models equipped with our strategy manifest performance metrics that rival, and at times surpass, those of considerably larger models. This not only confirms the efficacy of our rationale refinement strategy but also opens up a promising avenue where smaller, more efficient models can be rendered competitive in the multimodal reasoning landscape.

\section{Related Work}

\subsection{Chain-of-Thought}
% % introduce language-only CoT: CoT, CoT-SC, ToT, GoT, SoT
The Chain-of-Thought (CoT) paradigm has emerged as a transformative approach, aiming to elucidate the reasoning processes of large language models and boost their ability in a variety of NLP tasks. The vanilla CoT~\cite{cot} prompting strategy models to generate intermediate reasoning steps, leading to significant performance improvements across a spectrum of tasks. Its methodology laid the groundwork for subsequent advancements in the CoT paradigm. CoT-SC~\cite{cot-sc} introduced a self-consistency decoding strategy by sampling multiple reasoning paths and selecting the most consistent path to enhance the reliability of the generated rationales.

The versatility and adaptability of the CoT paradigm were further demonstrated with the advent of advanced reasoning structures. Tree-of-Thoughts (ToT)~\cite{tot} transitioned from linear chains to more intricate tree-like structures, aiming to capture more complex reasoning patterns. In a parallel development, \cite{got} ventured into graph-based reasoning, presenting the Graph-of-Thought (GoT) approach, which allowed for a more interconnected reasoning process. Skeleton-of-Thought (SoT)~\cite{sot} took a different route by emphasizing the benefits of efficiency by underscoring the potential of parallel decoding, enabling models to explore multiple reasoning paths simultaneously.

While the aforementioned studies have made significant strides in the CoT paradigm, we seek to extend these advancements into the realm of multimodal reasoning. Multimodal-CoT~\cite{mm-cot} is a seminal work, but the unsatisfying performance of its generated rationales has been a major bottleneck. Our work aims to fully explore the potential of the CoT paradigm in multimodal reasoning.

\subsection{Multimodal Visual Question Answering}

Multimodal Visual Question Answering (VQA) has emerged as a key research area bridging the domains of vision and language. Pioneering works like MCAN~\cite{mcan}, BAN~\cite{ban}, and Top-Down~\cite{topdown} have established a foundation by introducing diverse attention mechanisms. Building on this, DFAF~\cite{dfaf} introduced innovative dynamic fusion strategies to facilitate both intra- and inter-modality attention flows. The introduction of Transformer architectures has significantly impacted the VQA field, as demonstrated by state-of-the-art models such as ViLT~\cite{vilt} and VisualBERT~\cite{visualbert}. In the same vein, Patch-TRM~\cite{patchtrm} leveraged a pyramid cross-modal Transformer, paired with pre-trained input embeddings from the icon dataset, advancing the understanding of abstract diagrams. Recently, incorporating large language models into the multimodal framework has seen a surge, exemplified by models like LLaMA-Adapter~\cite{llama-adapter} and BLIP-2~\cite{blip2}. LaVIN~\cite{lavin} introduced a novel routing algorithm that helps the model autonomously switch between unimodal and multimodal instruction reasoning paths. These models employ frozen visual encoders paired with fine-tuned language models, thereby setting new standards in VQA tasks.

ScienceQA~\cite{scienceqa} introduced a reasoning process into the VQA task, setting a benchmark for multimodal chain-of-thought reasoning. MM-REACT~\cite{mmreact}, with its innovative prompting design, enables language models to process multimodal information, facilitating the integration of ChatGPT with visual experts. A-OKVQA~\cite{a-okvqa} requires commonsense reasoning about the depicted scene in an image. Multimodal-CoT~\cite{mm-cot} integrates language (text) and vision (image) modalities into a two-stage framework, distinctly separating rationale generation from answer inference.

In this work, we typically focus on multimodal reasoning tasks, a subtype of multimodal visual question answering (VQA) that incorporates rationales. The significance of this task lies in its ability to more accurately simulate human cognitive processes. When humans answer questions, they often rely on observed visual information and existing textual knowledge, synthesizing the two to form a reasoned conclusion. These multimodal reasoning tasks comprehensively assess a model's capacity to understand and fuse visual and textual information, thereby propelling the development of multimodal learning. Moreover, these tasks enhance the interpretability of artificial intelligence systems. Because models must supply not only answers but also supporting rationales, their decision-making processes become more interpretable and comprehensible.

\section{Method}

\subsection{Preliminaries}

In this work, we focus on multimodal reasoning, a challenging task that involves reasoning across multiple modalities to answer questions. Specifically, we consider visual question answering~\cite{antol2015vqa,wu2018chain}, where the model is tasked with answering questions based on the information provided in the question itself and the accompanying image. 

Formally, given a dataset $\mathcal{D} = \{\mathcal{X}, \mathcal{Y}\}$, where $X \in \mathcal{X}$ represents a multimodal input consisting of text and images, and $Y \in \mathcal{Y}$ represents the corresponding output, the task of multimodal reasoning involves learning a mapping function parameterized by $\Theta$: $\mathcal{F}_\Theta: \mathcal{X} \rightarrow \mathcal{Y}$ that accurately predicts the output $Y$ for a given input $X$. The multimodal input $X$ can be represented as $X = (T, I)$, where $T$ is the text component, and $I$ is the image component of the input. The function $\mathcal{F}$ should be able to effectively leverage the information from both the text and image modalities to make accurate predictions. Specifically, the basic visual question answering task is defined as:
\begin{equation}
\begin{aligned}
    Y = \arg\max_{Y'} p(Y' \;| \; T, \; I)
\end{aligned}
\end{equation}
where $p(Y' \;| \; T, \; I)$ is the probability of answer $Y'$ given the text $T$ and image $I$.

The multimodal reasoning task builds upon the basic visual question answering task, extending its requirements to necessitate the generation of a rationale $R'$ that elucidates the reasoning process underpinning the answer $Y$. This task can be mathematically represented as follows:
\begin{equation}
    Y, R = \arg\max_{Y', R'} p(Y', R' \;| \; T, \; I)
\end{equation}

\paragraph{Multimodal-CoT}~\cite{mm-cot} decomposed the multimodal reasoning task into a two-stage framework, where the model first generates a rationale $R$ and then utilizes this rationale to predict the answer $Y$. The process is delineated as:
\begin{equation}
\begin{aligned}
    R &= \arg\max_{R'} p(R' \; | \; T, \; I) \\
    Y &= \arg\max_{Y'} p(Y' \; | R,  \; T, \; I)
\end{aligned}
\end{equation}
In this work, we follow the same two-stage framework as Multimodal-CoT~\cite{mm-cot} and explore the potential of such a chain-of-thought reasoning paradigm.

\subsection{Multimodal Consistent Chain-of-Thought}

Though the existing multimodal CoT framework has been proven to be beneficial in multimodal reasoning, there remains a large room to fully explore its potential. In this work, we introduce a simple yet effective training strategy, \textbf{M}ultimodal \textbf{C}onsistent \textbf{C}hain-of-\textbf{T}hought (MC-CoT), to learn high-quality rationales and thus boost the reasoning performance of the multimodal CoT framework.

\begin{figure*}[t]
\centering
\includegraphics[width=1.0\textwidth]{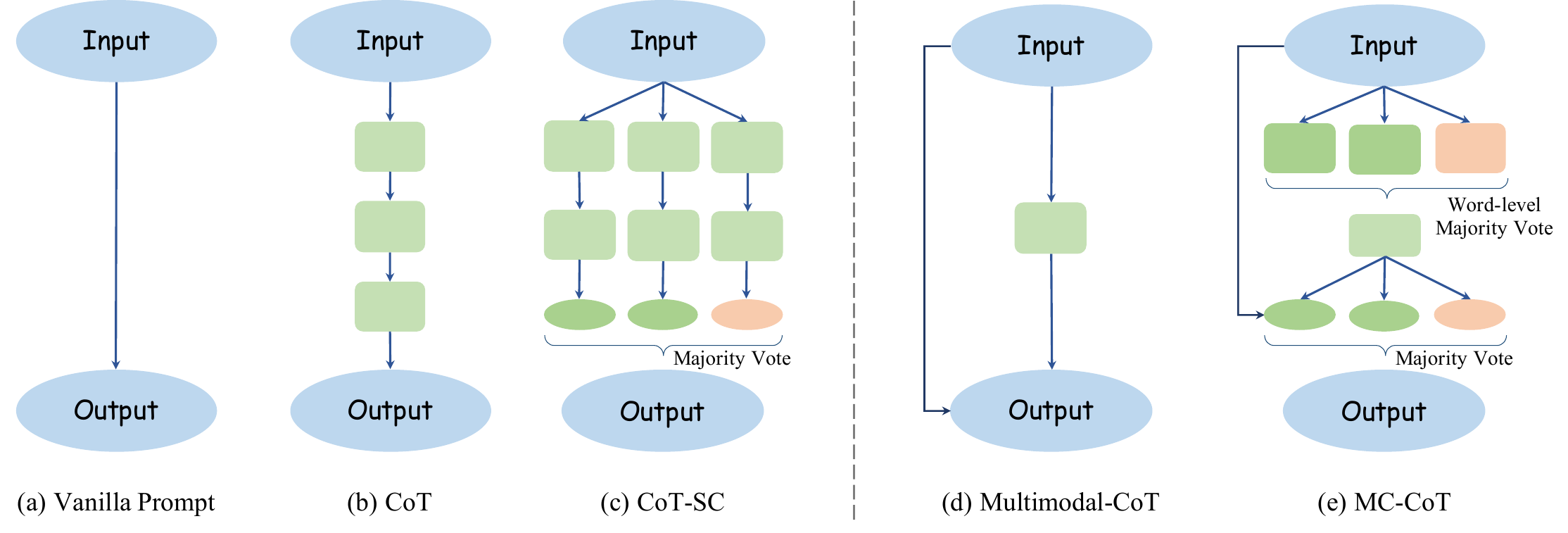}
\caption{A comparison schematic diagram of different Chain-of-Thought (CoT) prompt-based reasoning methods. (a) The basic input-output prompt, (b) Chain-of-Thought with intermediate chain-like reasoning, (c) Chain-of-Thought Self-Consistency (CoT-SC) that utilizes the consistency of multiple independent chains of thoughts for reasoning, (d) Multimodal-CoT, which infers the rationale using the input text and image, and then predicts the answer using the rationale as part of the input, and (e) MC-CoT that infers a high-quality rationale through word-level voting, and then obtains a high-quality answer using majority vote. \textit{It is worth noting that our approach leverages multiple chain consistency only during the training phase, in contrast to CoT-SC, which employs it during the inference stage.}}
\label{fig:framework_comparison}
\end{figure*}

\paragraph{Rationale Generation} Leveraging the inherent randomness introduced by the dropout operations, the model is able to generate multiple diverse rationales for a given input text-image pair $(T, I)$ by sampling from the model $N_r$ times:
\begin{equation}
    R^i \sim p(R \; | \; T, \; I), \; i = 1, 2, \dots, N_r
\end{equation}
We perform a voting process on the generated rationales to select the most consistent words across the rationales. Specifically, the best rationale $R^*$ is selected as follows:
\begin{equation}
    R^*_j = \text{Vote}(\{R^i_j\}_1^{N_r}),
\end{equation}
where, for each position $j$ in the rationale, we choose the majority of the words at position $j$ across the $N_r$ generated rationales to form the best rationale $R^*_j$. 

\paragraph{Answer Inference} In the answer inference stage, we also generate multiple answers based on the best rationale $R^*$ and the input text-image pair $(T, I)$ by sampling from the model $N_a$ times:
\begin{equation}
    Y^i \sim p(Y \; | \; R^*, \; T, \; I), \; i = 1, 2, \dots, N_a.
\end{equation}
The final answer is selected by the majority voting:
\begin{equation}
    Y^* = \text{Vote}(\{Y^i\}_1^{N_a}).
\end{equation}
where the majority of the answers across the $N_a$ generated answers are regarded as the most convincing answers $Y^*$. 
The pseudocode of the pipeline is shown in Appendix~\ref{app:pseudocode}.

\paragraph{Voting Strategy} Consider the answer inference stage as an illustration. The vanilla Multimodal-CoT model computes a cross-entropy loss between a $d$-dimensional logit $L \in \mathbb{R}^d$ and the the ground truth label $Y \in \mathbb{R}^d$, which can be represented as:
\begin{equation}
    \mathcal{L} = \text{CrossEntropy}(L, Y),
\end{equation}
Our averaging-based voting strategy initially calculates the mean logit $\bar{L} \in \mathbb{R}^d$ and weighted mean logit $\hat{L} \in \mathbb{R}^d$:
\begin{equation}
\begin{aligned}
\bar{L} = \frac{1}{N_a} \sum_{i=1}^{N_a} L^i, \; \;
\hat{L} = \frac{1}{N_a} \frac{\sum_{i=1}^{N_a} w L^i}{\sum_j w_j},
\end{aligned}
\end{equation}
where $w = 1 / (1 + \sigma) \in \mathbb{R}^d$ is a weight vector derived from the standard deviation of the logits, calculated across different predictions for each dimension. The weight for each logit's dimension is inversely proportional to its variability, assigning higher confidence to predictions with lower variability and lower confidence to those with higher variability. The final prediction is then a linear combination of $\bar{L}$ and $\hat{L}$:
\begin{equation}
\begin{aligned}
    L^* &= \alpha \bar{L} + (1-\alpha) \hat{L} \\
\end{aligned}
\label{eq:final_logit}
\end{equation}
with $\alpha$ being a hyperparameter set empirically to $0.5$. The cross-entropy loss is subsequently computed between 
$L^*$ and the ground truth label $Y$:
\begin{equation}
\mathcal{L} = \text{CrossEntropy}(L^*, Y).
\end{equation}

We present a comparison of our approach with existing CoT-based reasoning methods in Fig.~\ref{fig:framework_comparison}. Specifically, subfigures (a-c) illustrate the prompt learning methods commonly employed for text modality in large language models, while subfigures (d-e) depict the CoT reasoning frameworks in the multimodal field. It can be seen that our approach shares similarities with CoT-SC, such as the utilization of a voting mechanism. However, there are also significant differences between the two approaches: (i) the objective differs. CoT-SC aims to facilitate prompt learning in large language models during the inference stage, while our method seeks to enhance the robustness of model reasoning during the training phase; (ii) the voting process differs. CoT-SC only votes on the final answers inferred from each thought, whereas our method incorporates voting during the rationale generation stage.

\section{Theoretical Insights}

\subsection{Aggregation Minimizes Expected Loss}

\noindent{\textbf{Theorem 1.}} (Jensen's inequality) Let $X$ be a random variable and let $\phi$ be a convex function. Then the following inequality holds: $\phi(\mathbb{E}[X]) \leq \mathbb{E}[\phi(X)]$.

Our proposed method, MC-CoT, leverages the variability introduced by dropout techniques to generate a diverse set of rationales. By aggregating these rationales, we aim to produce explanations and corresponding answers of higher quality. In the context of our framework, the convex function $\phi$ is represented by the cross-entropy loss function, and the random variable in question is the set of logits $L^i$. By applying Jensen's inequality to our model, we deduce that the aggregation process leads to a lower expected loss: $\mathcal{L}(\mathbb{E}(L^*), Y) \leq \mathbb{E}(\mathcal{L}(L^*, Y))$.

\subsection{Bias-Variance Trade-off}
\vspace{-1mm}

We introduce a linear combination of two voting strategies to derive the final prediction, as delineated in Equation~\ref{eq:final_logit}. We can analytically decompose the expected value of the squared difference between the predicted outcomes $Y^*$ and the ground truth $Y$ into three distinct components:
\begin{equation}
    \mathbb{E}[(Y^* - Y)^2] = \textrm{Bias}^2(Y^*) + \textrm{Var}(Y^*) + \epsilon,
\end{equation}
where $\textrm{Bias}^2(Y^*) = (\mathbb{E}[Y^*] - Y)^2$ quantifies the systematic deviation of the expected prediction from the ground truth, indicative of the error inherently introduced by the predictive model itself. Meanwhile, $\textrm{Var}(Y^*) = \mathbb{E}[(Y^* - \mathbb{E}[Y^*])^2]$ captures the variability of the predictions, reflecting the extent to which these predictions will fluctuate around their expected value. Lastly, $\epsilon$ denotes the irreducible error, encapsulating the intrinsic noise within the data.

\textit{The mean logits}, denoted by $\bar{L}$, may exhibit lower bias since they represent the composite prediction averaged across multiple models. Employing solely the mean logits promotes exploration by treating all predictions as equally informative. Nevertheless, its measure of central tendency does not account for the individual variability within the predictions. \textit{The weighted mean logits}, denoted by $\hat{L}$, aim to reduce variability by assigning greater significance to consistent (low-variance) predictions. This reflects a confidence-based strategy, allocating higher weight to predictions that are less variable. However, it could introduce bias if predictions are consistently erroneous.

The final logits, which include both mean logits and weighted mean logits, are designed to achieve an optimal balance between bias and variance. While the mean logits promote exploration by treating all predictions equally, the weighted mean logits perform exploitation by assigning higher weights to consistent predictions. The combination allows balancing exploration with exploitation, thus improving the performance towards an optimal standard.

\section{Experiment}

\paragraph{Datasets} We evaluated MC-CoT on two prominent multimodal reasoning benchmarks. The first, ScienceQA~\cite{scienceqa}, is a comprehensive multimodal science question dataset that contains annotated answers with detailed reasoning and explanations. It encompasses more than 21,000 multiple-choice questions, showcasing a vast range of domains that span three subjects, include 26 topics, cover 127 categories, and incorporate 379 distinct skills. The benchmark dataset is divided into training, validation, and test sets, consisting of 12,726, 4,241, and 4,241 examples, respectively. The second dataset is A-OKVQA~\cite{a-okvqa}, a knowledge-based visual question answering dataset that contains 25,000 questions requiring a broad base of commonsense and world knowledge for accurate responses, divided into 17,000 training, 1,000 validation, and 7,000 test examples.

\paragraph{Baselines}
In our assessment of the ScienceQA dataset, we have compared MC-CoT against recent strong baselines, covering five categories of methodologies: (i) heuristic and expert-guided choices, such as random choice and human evaluation~\cite{scienceqa}; (ii) standard multimodal visual question answering approaches, which include MCAN~\cite{mcan}, Top-Down~\cite{topdown}, BAN~\cite{ban}, DFAF~\cite{dfaf}, ViLT~\cite{vilt}, Patch-TRM~\cite{patchtrm}, and VIsualBERT~\cite{visualbert}; (iii) Instruction-tuned large language models like GPT-3.5~\cite{gpt35} and its CoT-enhanced variants~\cite{scienceqa}, in addition to ChatGPT, GPT-4~\cite{gpt-4}, and Chameleon~\cite{chameleon}; (iv) Specifically finetuned large language models, notably LLaMA-Adapter~\cite{llamaadapter}, LaVIN~\cite{lavin}, LLaMA-SciTune~\cite{scitune}, and LLaVa~\cite{llava}; (v) multimodal reasoning models based on chain-of-thought, Multimodal-CoT~\cite{mm-cot} and our MC-CoT both belong to this category. Regarding the A-OKVQA dataset, the baselines encompass key methods that have propelled the field of visual question answering forward, such as Pythia~\cite{Pythia}, ViLBERT~\cite{vilbert}, LXMERT~\cite{lxmert}, KRISP~\cite{krisp}, GPV-2~\cite{gpv2}, BLIP-2~\cite{blip2}, PICa~\cite{PICA}, and IPVR~\cite{IPVR}. Additionally, Multimodal-CoT \cite{mm-cot} is incorporated as the chain-of-thought multimodal benchmark for this dataset.

\paragraph{Implementation Details}

Following the experimental setup detailed in \cite{mm-cot}, we implemented the T5 encoder-decoder framework \cite{raffel2020exploring}. We utilize the UnifiedQA~\cite{UNIFIEDQA} models, which have 223M and 738M parameters, as our default Base and Large models, respectively. Additionally, we employ the FLAN-T5\cite{FlanT5XXL} models with 248M and 783M parameters as our F-Base and F-Large models. Batch sizes were set to $16$ for the base model and $8$ for the large model. The learning rate was set to \(5 \times 10^{-5}\). Regarding sequence length, the initial output was constrained to $512$ tokens, while the subsequent output was limited to $64$ tokens. The experiments utilized four NVIDIA Tesla A100 GPUs, each equipped with 80GB of memory.

\begin{table}[!h]
\centering
\caption{The comparison on ScienceQA dataset. Question classes: NAT = natural science, SOC = social science, LAN = language science, TXT = text context, IMG = image context, NO = no context, G1-6 = grades 1-6, G7-12 = grades 7-12.}
{\renewcommand\baselinestretch{1.1}
\selectfont
\resizebox{\textwidth}{!}{
\begin{tabular}{ccccccccccc}
\toprule
Model & Size & NAT & SOC & LAN & TXT & IMG & NO & G1-6 & G7-12 & AVG \\
\midrule
Random Choice~\cite{scienceqa} & - & 40.28 & 46.13 & 29.25 & 47.75 & 40.08 & 33.66 & 39.35 & 40.67 & 39.83 \\
Human~\cite{scienceqa} & - & 90.23 & 84.97 & 87.48 & 89.60 & 87.50 & 88.10 & 91.59 & 82.42 & 88.40 \\
\hline
MCAN~\cite{mcan} & 95M & 56.08 & 46.23 & 58.09 & 59.43 & 51.17 & 55.40 & 51.65 & 59.72 & 54.54 \\
Top-Down~\cite{topdown} & 70M & 59.50 & 54.33 & 61.82 & 62.90 & 54.88 & 59.79 & 57.27 & 62.16 & 59.02 \\
BAN~\cite{ban} & 112M & 60.88 & 46.57 & 66.64 & 62.61 & 52.60 & 65.51 & 56.83 & 63.94 & 59.37 \\
DFAF~\cite{dfaf} & 74M & 64.03 & 48.82 & 63.55 & 65.88 & 54.49 & 64.11 & 57.12 & 67.17 & 60.72 \\
ViLT~\cite{vilt} & 113M & 60.48 & 63.89 & 60.27 & 63.20 & 61.38 & 57.00 & 60.72 & 61.90 & 61.14 \\
Patch-TRM~\cite{patchtrm} & 90M & 65.19 & 46.79 & 65.55 & 66.96 & 55.28 & 64.95 & 58.04 & 67.50 & 61.42 \\
VisualBERT~\cite{visualbert} & 111M & 59.33 & 69.18 & 61.18 & 62.71 & 62.17 & 58.54 & 62.96 & 59.92 & 61.87 \\
UnifiedQA$_{\text{Base}}$~\cite{UNIFIEDQA} & 223M & 68.16 & 69.18 & 74.91 & 63.78 & 61.38 & 77.84 & 72.98 & 65.00 & 70.12 \\
UnifiedQA$_{\text{Base}}$ w/ CoT~\cite{scienceqa} & 223M & 71.00 & 76.04 & 78.91 & 66.42 & 66.53 & 81.81 & 77.06 & 68.82 & 74.11 \\
\hline
GPT-3.5~\cite{scienceqa} & 173B & 74.64 & 69.74 & 76.00 & 74.44 & 67.28 & 77.42 & 76.80 & 68.89 & 73.97 \\
GPT-3.5 w/ CoT~\cite{scienceqa} & 173B & 75.44 & 70.87 & 78.09 & 74.68 & 67.43 & 79.93 & 78.23 & 69.68 & 75.17 \\
ChatGPT w/ CoT~\cite{gpt-4} & - & 78.82 & 70.98 & 83.18 & 77.37 & 67.92 & 86.13 & 80.72 & 74.03 & 78.31 \\
GPT-4 w/ CoT~\cite{gpt-4} & - & 85.48 & 72.44 & 90.27 & 82.65 & 71.49 & 92.89 & 86.66 & 79.04 & 83.99 \\
Chameleon + ChatGPT~\cite{chameleon} & - & 81.62 & 70.64 & 84.00 & 79.77 & 70.80 & 86.62 & 81.86 & 76.53 & 79.93 \\
Chameleon + GPT-4~\cite{chameleon} & - & 89.83 & 74.13 & 89.82 & 88.27 & 77.64 & 92.13 & 88.03 & 83.72 & 86.54 \\
\hline
LLaMA-Adapter (T)~\cite{llamaadapter} & 6B & 79.00 & 73.79 & 80.55 & 78.30 & 70.35 & 83.14 & 79.77 & 75.68 & 78.31 \\
LLaMA-Adapter~\cite{llamaadapter} & 6B & 84.37 & 88.30 & 84.36 & 83.72 & 80.32 & 86.90 & 85.83 & 84.05 & 85.19 \\
LaVIN-7B~\cite{lavin} & 7B & 89.25 & 94.94 & 85.24 & 88.51 & 87.46 & 88.08 & 90.16 & 88.07 & 89.41 \\
LLaMA-SciTune$_{\text{Base}}$~\cite{scitune} & 7B & 84.50 & 94.15 & 82.91 & 88.35 & 83.64 & 88.74 & 85.05 & 85.60 & 86.11 \\
LaVIN-13B~\cite{lavin} & 13B & 90.32 & 94.38 & 87.73 & 89.44 & 87.65 & 90.31 & 91.19 & 89.26 & 90.50 \\
LLaVa~\cite{llava} & 13B & 90.36 & 95.95 & 88.00 & 89.49 & 88.00 & 90.66 & 90.93 & 90.90 & 90.92 \\
LLaVa + GPT-4~\cite{llava} & 13B & 91.56 & \textbf{96.74} & 91.09 & 90.62 & 88.99 & 93.52 & 92.73 & 92.16 & 92.53 \\
LLaMA-SciTune$_{\text{Large}}$~\cite{scitune} & 13B & 89.30 & 95.61 & 87.00 & 93.08 & 86.67 & 91.75 & 84.37 & 91.30 & 90.03 \\
\hline
Mutimodal-CoT$_{\text{Base}}$~\cite{mm-cot} & 223M & 87.52 & 77.17 & 85.82 & 87.88 & 82.90 & 86.83 & 84.65 & 85.37 & 84.91 \\
Mutimodal-CoT$_{\text{Large}}$~\cite{mm-cot} & 738M & 95.91 & 82.00 & 90.82 & 95.26 & 88.80 & 92.89 & 92.44 & 90.31 & 91.68 \\
% \rowcolor[HTML]{E7ECE4} 
MC-CoT$_{\text{Base}}$ & 223M & 91.87 & 84.59 & 93.00 & 92.28 & 88.30 & 92.75 & 90.64 & 90.64 & 90.64 \\
% \rowcolor[HTML]{E7ECE4} 
MC-CoT$_{\text{Large}}$ & 738M & 95.47  & 89.99 &  91.82 & 95.11 & 92.66 & 93.24  & 94.27 & 91.76 & 93.37  \\
% \rowcolor[HTML]{E7ECE4} 
MC-CoT$_{\text{F-Base}}$ & 248M & 93.56 & 83.58 & 90.73 & 94.13  & 89.24  &  90.94 & 90.93 & 90.38 & 90.73  \\
% \rowcolor[HTML]{E7ECE4} 
MC-CoT$_{\text{F-Large}}$ & 783M & \textbf{97.47} & 90.44 & \textbf{93.18} & \textbf{96.97} & \textbf{93.75} & \textbf{94.49} & \textbf{95.30} & \textbf{94.13} & \textbf{94.88} \\
\bottomrule
\end{tabular}
} \par}
\label{tab:scienceqa}
\end{table}
    
\subsection{Main Results}

We conducted a comparison of MC-CoT with state-of-the-art approaches on the ScienceQA dataset, as depicted in Table~\ref{tab:scienceqa}. We present the overall model size rather than the tunable model size, as it more accurately reflects the model's capacity. The results indicate that our MC-CoT$_{\text{Base}}$ model, with a model size of only 223 million parameters, achieved an average accuracy of $90.64\%$, approaching the performance of Lavin-13B and LLaVa-13B models that were fine-tuned based on larger language models. Furthermore, our MC-CoT$_{\text{F-Large}}$ model, with 783 million parameters, achieved state-of-the-art results, surpassing the strongest fine-tuned large language model baseline LLaVa+GPT-4 by $2.35\%$. 

Furthermore, in comparison to multimodal chain-of-thought baselines, our MC-CoT$_{\text{Base}}$ model showcased an average accuracy improvement of $5.70\%$ compared to Multimodal-CoT$_{\text{Base}}$. Similarly, our MC-CoT$_{\text{Large}}$ model exhibited an improvement of $1.69\%$ compared to the Multimodal-CoT Large model. These observations demonstrate that our approach has the potential to significantly enhance the model's reasoning capabilities. 

We report the experimental results on the A-OKVQA dataset within Table~\ref{tab:aokvqa_val}. Our MC-CoT$_{\text{Base}}$ model demonstrates exceptional improvements over the current state-of-the-art models, both in terms of direct-answer accuracy and multi-choice task performance. Specifically, MC-CoT$_{\text{Base}}$ achieves a direct-answer accuracy of $68.7\%$, substantially outperforming the strongest competitor, BLIP-2, which possesses an extensive parameter count of over 11 billion. This represents an improvement of over $15.5\%$, a margin that underscores the efficiency and effectiveness of our model's reasoning capabilities. In the multi-choice task, MC-CoT$_{\text{Base}}$ reaches an accuracy of $71.0\%$, which not only exceeds BLIP-2's performance by $0.8\%$ but also does so with significantly fewer parameters. This comparison highlights the fact that our model's parameter utilization is exceptionally efficient, leading to enhanced performance even against models with far greater complexity.

\begin{table*}[ht]
\centering
\caption{The comparison on A-OKVQA dataset. We conduct the evaluation on both direct-answer and multi-choice tasks.}
{\renewcommand\baselinestretch{1.1}\selectfont
\resizebox{1.0\textwidth}{!}{
\begin{tabular}{cccccc}
\toprule
Model & Vision Model & Text Model & Parameters & Direct-answer & Multi-choice \\
\midrule
Pythia~\cite{Pythia} & ResNet~\cite{resnet} & BERT~\cite{bert} & 70M & 25.2 & 49.0 \\
ViLBERT~\cite{vilbert} & Faster R-CNN~\cite{fastrcnn} & BERT~\cite{bert} & 300M & 30.6 & 49.1 \\
LXMERT~\cite{lxmert} & Transformer~\cite{attention} & Transformer~\cite{attention} & 220M & 30.7 & 51.4 \\
KRISP~\cite{krisp} & Faster R-CNN~\cite{fastrcnn} & BERT~\cite{bert} & 200M & 33.7 & 51.9 \\
GPV-2~\cite{gpv2} & VinVL~\cite{vinvl} & T5-Base~\cite{t5} & 300M & 48.6 & 60.3 \\
\hline
BLIP-2~\cite{blip2} & CLIP-VIT-LARGE~\cite{clip} & FlanT5XXL~\cite{FlanT5XXL} & 11B & \underline{53.2} & \underline{70.2} \\
PaLM-CoT~\cite{cot} & - & PaLM~\cite{palm} & 540B &41.5  & 48.1 \\
PICa~\cite{PICA}& VinVL~\cite{vinvl} & GPT-3~\cite{gpt3} & 175B & 42.4 & 46.1 \\
IPVR~\cite{IPVR}& Faster-RCNN~\cite{fastrcnn} & OPT~\cite{opt} & 66B  & 46.4 & 48.6 \\
\hline
Mutimodal-CoT$_{\text{Base}}$~\cite{mm-cot} & DETR~\cite{detr} & UnifiedQABase~\cite{UNIFIEDQA} & 223M & - & 50.6 \\
% \rowcolor[HTML]{E7ECE4} 
MC-CoT$_{\text{Base}}$ & DETR~\cite{detr} & UnifiedQABase~\cite{UNIFIEDQA} & 223M & \textbf{68.7} & \textbf{71.0} \\
% \rowcolor[HTML]{E7ECE4} 
% Ours$_{\text{Large}}$ & detr~\cite{detr} & FLAN-T5Large~\cite{FlanT5XXL} & 738M & \textbf{72.1} & \textbf{78.8} \\
\bottomrule
\end{tabular}} \par}
\label{tab:aokvqa_val}
\end{table*}

These results are particularly noteworthy given the complexity and diversity of the A-OKVQA dataset, which is designed to challenge models with questions that require advanced understanding and reasoning across both textual and visual modalities. The performance leap made by MC-CoT$_{\text{Base}}$ reinforces the potential of our improved training strategy. By introducing a voting mechanism among multiple rationales, our approach refines the quality of the generated rationales, which in turn significantly enhances the accuracy of the final answer. This innovative training paradigm proves effective in a multimodal context, where the interplay between textual and visual elements is critical for achieving high performance. Moreover, the robustness of MC-CoT$_{\text{Base}}$ suggests that it is not only the size of the model that dictates performance but also the methodological advancements that come with thoughtful design and algorithmic improvements. This is evident from the fact that our base-sized model can outstrip the performance of models with parameter counts orders of magnitude larger.

\subsection{Quantative Analysis} 
    
\subsubsection{Ablation Study}

We present the experimental results on the ScienceQA dataset illustrated in Table~\ref{tab:ablation}. It can be seen that both using $\hat{Y}$ only and $\bar{Y}$ only can achieve comparable performance but the combined model outperforms them. This indicates that the two voting strategies are complementary to each other. Moreover, we can see that the model without voting R (rationales) obtains a significant drop in performance across all categories, with the average accuracy plummeting to 84.70\%. This highlights the importance of the rationale voting in improving the model's reasoning capabilities and accuracy. The absence of the answer voting mechanism results in a decrease in performance in the SOC, LAN, NO, G1-6, and G7-12 categories, with a slight improvement in NAT and TXT. The overall average is 90.19\%, indicating that answer voting contributes to the overall robustness, albeit not as critically as rationale voting. Inference voting means that voting is only conducted during the inference stage and not during the training stage. The significant performance drop of this strategy indicates that inference phase voting is not as effective as training phase voting.

\begin{table*}[h]
\vspace{-4mm}
\centering
\caption{Ablation study on ScienceQA dataset.}
\setlength{\tabcolsep}{1.4mm}{
\begin{tabular}{cccccccccc}
\toprule
Model & NAT & SOC & LAN & TXT & IMG & NO & G1-6 & G7-12 & AVG \\
\midrule
MC-CoT$_{\text{Base}}$ & 91.87 & 84.59 & 93.00 & 92.28 & 88.30 & 92.75 & 90.64 & 90.64 & 90.64 \\
\hline
w/ mean only      & 92.10 & 83.80 & 90.82 & 92.47 & 88.84 & 90.66 & 90.16 & 89.78 & 90.03 \\
w/ weighted only  & 91.61 & 83.69 & 91.00 & 92.38 & 88.45 & 90.80 & 89.72 & 89.91 & 89.79 \\
w/o voting R      & 87.34 & 77.39 & 85.18 & 87.68 & 83.24 & 86.34 & 84.58 & 84.90 & 84.70 \\
w/o voting A      & 92.36 & 84.14 & 90.64 & 92.42 & 88.30 & 91.50 & 90.35 & 89.91 & 90.19 \\
inference voting  & 87.43 & 77.28 & 84.64 & 87.39 & 82.85 & 86.41 & 84.91 & 83.98 & 84.58 \\
\bottomrule
\end{tabular}
}
\label{tab:ablation}
\vspace{-6mm}
\end{table*}

\subsubsection{Effect of Rationale Generation}

\begin{figure}[ht]
  \centering
  \begin{minipage}[h]{0.48\linewidth} % Adjust the minipage width to fit your content
      \includegraphics[width=\textwidth]{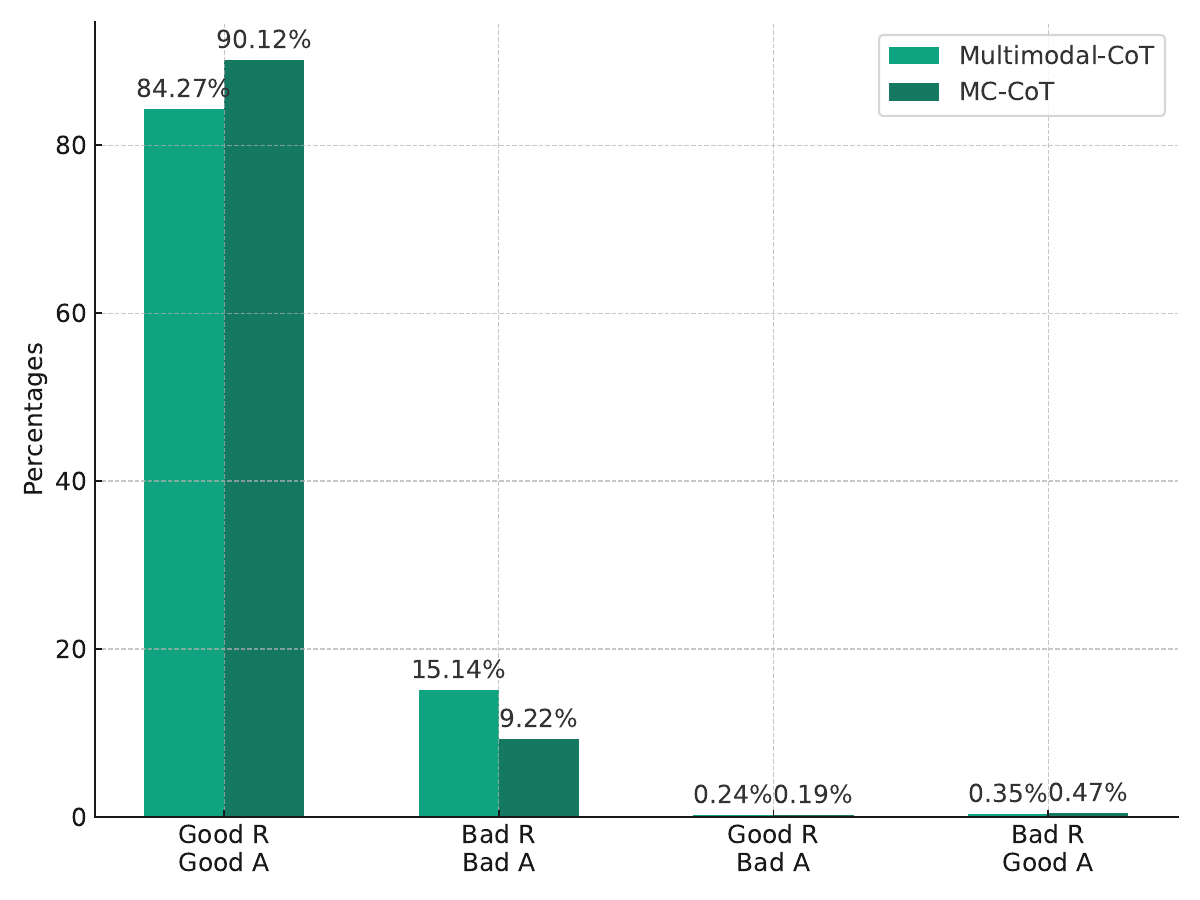}
      \caption{The relationship between rationale and answer.}
      \label{fig:rationale_vs_answer}
  \end{minipage}
  \hfill % Adds horizontal space between the figure and table
  \begin{minipage}[h]{0.48\linewidth} % Adjust the minipage width to fit your content
      \captionof{table}{The RougeL score of the generated rationales and the average accuracy of predicted answers on the ScienceQA dataset.} % Use captionof for table caption
      \setlength{\tabcolsep}{0.75mm}{
      \begin{tabular}{ccc}
          \toprule
          Method & RougeL & Avg Acc \\
          \midrule
          Multimodal-CoT$_{\textrm{Base}}$   & 96.97 & 84.91  \\
          MC-CoT$_{\textrm{Base}}$           & 97.98 & 90.64  \\
          MC-CoT$_{\textrm{F-Large}}$        & 98.47 & 94.88  \\
          \bottomrule
      \end{tabular}}
      \label{tab:effect}
  \end{minipage}
\end{figure}

We categorized the predictions from the Multimodal-CoT$_{\textrm{Base}}$ and MC-CoT$_{\textrm{Base}}$ models into four types: (i) Good Rationale and Answer; (ii) Bad Rationale and Answer; (iii) Good Rationale but Bad Answer; (iv) Bad Rationale but Good Answer. A 'Good Rationale' indicates the high-quality generated rationale. The results are depicted in Fig.~\ref{fig:rationale_vs_answer}. It can be seen that MC-CoT exhibits a higher frequency of Good Rationale and Answer, and a lower frequency of Bad Rationale and Answer. This suggests the self-consistency training approach improves the rationale's quality and the accuracy of the answers.

We present the evaluation of the generated rationales and answers on the ScienceQA dataset in Table~\ref{tab:effect}. It can be seen that our proposed MC-CoT$_{\textrm{Base}}$ model generated rationales that improved RougeL scores by approximately 1\% compared to the Multimodal-CoT$_{\textrm{Base}}$ model. However, this led to an approximately 6\% increase in the average accuracy of predicted answers. Similarly, the rationales generated by the MC-CoT$_{\textrm{F-Large}}$ model showed only about a 0.5\% improvement in RougeL scores compared to the MC-CoT base model. Yet, this resulted in an approximate 4\% increase in the average accuracy of predicted answers. This demonstrates that even slight improvements in the quality of rationales can impact the answer inference of the multimodal reasoning models.

\subsection{Qualitative Analysis}

\begin{figure}[!h]
\centering
\includegraphics[width=0.98\textwidth]{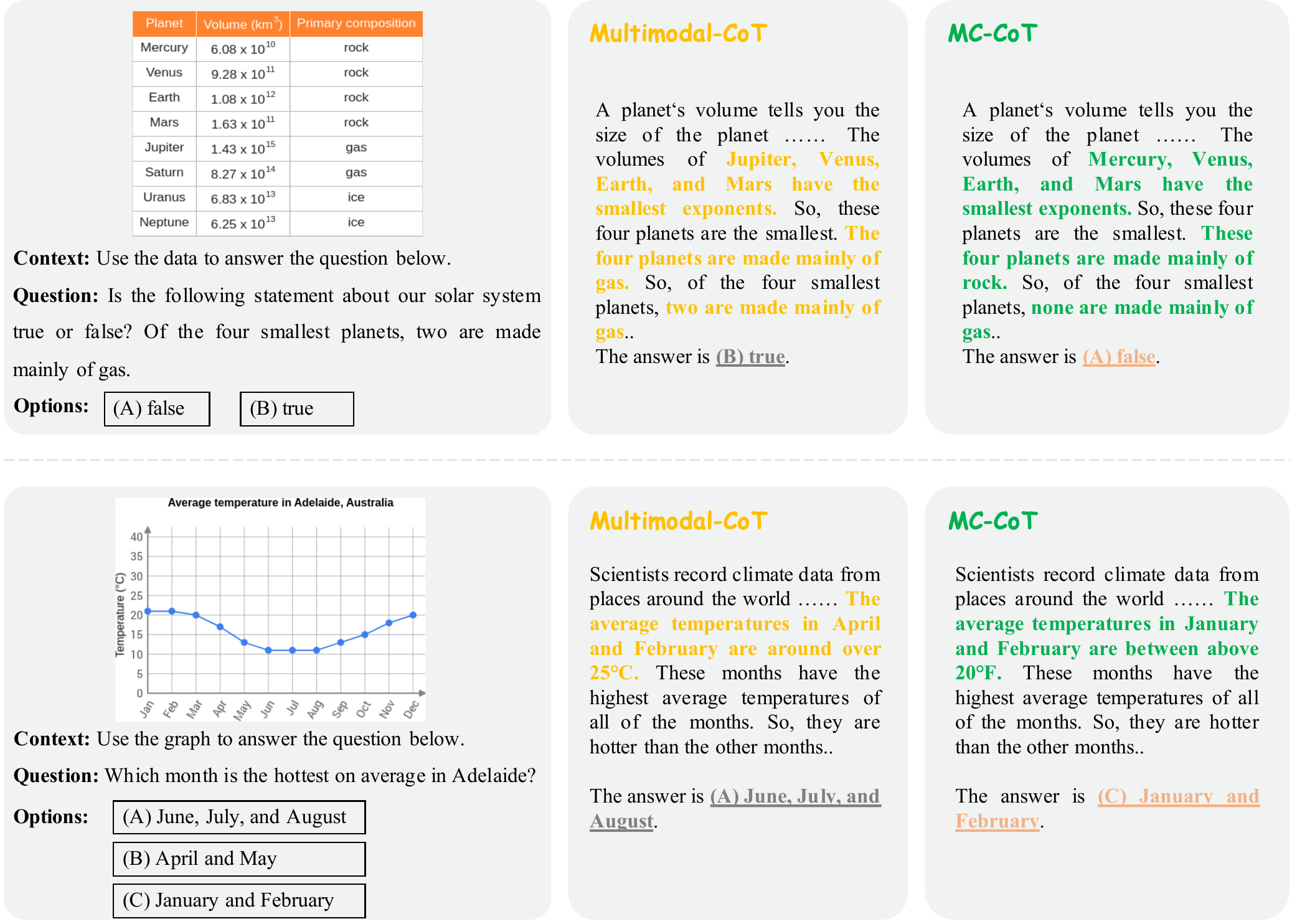}
\caption{Comparison on predicted examples.}
\label{fig:predict_example}
\end{figure}

We present two illustrative examples of predictions, as depicted in Figure~\ref{fig:predict_example}, to demonstrate the efficacy of our proposed model. These examples have been carefully selected to showcase the model's capabilities in handling complex patterns and making accurate choices. For readers interested in exploring further, a more detailed collection of examples is provided in Appendix~\ref{app:examples}. 

The first example highlights a scenario where the Multimodal-CoT model encountered difficulties in generating a coherent and accurate rationale, ultimately leading to an erroneous conclusion. This misstep underscores the challenges inherent in integrating multiple sources of information and extracting a logical reasoning path. Conversely, the MC-CoT model demonstrated its proficiency by navigating these complexities successfully, thereby arriving at the correct answer. This contrast exemplifies the significance of robust reasoning capabilities in achieving high accuracy in complex question-answering tasks.

The second example delves into a different set of challenges faced by both models. Here, the Multimodal-CoT model stumbled over the interpretation of data presented in a chart, a mistake that significantly derailed its reasoning process and resulted in an incorrect answer. This error highlights the critical importance of accurate data interpretation skills in models designed for multimodal contexts. On the other hand, the MC-CoT model, while generally more adept in this instance, exhibited a lapse in understanding units of measurement—mistaking degrees Celsius ($^\circ\text{C}$) for degrees Fahrenheit ($^\circ\text{F}$). Interestingly, this error did not compromise the integrity of its final answer, suggesting a certain resilience in the model's reasoning process despite minor inaccuracies.

\section{Conclusion}

In this study, we conducted an exhaustive analysis of the Multimodal-CoT framework and identified that the quality of rationales plays a crucial role in enhancing the performance of multimodal reasoning models. Recognizing the critical importance of this aspect, we developed an innovative solution aimed at improving the quality of rationales. Our solution employs a self-consistency training strategy that leverages the natural variability introduced by dropout techniques during training to foster a consensus mechanism through voting. This approach is supported by both theoretical underpinnings and empirical evidence, highlighting its effectiveness and wide applicability. We meticulously designed experiments to evaluate the impact of our self-consistency training strategy on the performance of multimodal reasoning models across diverse scenarios and datasets. The results from these experiments were revealing, clearly showing that our method significantly boosts the capabilities of multimodal models. A particularly significant finding was our approach's ability to enhance the performance of smaller models, enabling them to match or even exceed the capabilities of much larger models. We believe our strategy offers a promising avenue for advancing multimodal reasoning.

A limitation of our approach is that we do not explicitly impose grammatical constraints, meaning the generated rationales are not guaranteed to be grammatically correct. It is worth noting that the Multimodal-CoT framework also does not guarantee the grammatical validity of the generated rationales. Despite this, empirical evidence has shown that our approach can yield superior results across various datasets, indicating that the rationales produced by our approach can significantly enhance the model's reasoning ability towards the correct answers.

\section{Acknowledgement}

This work was supported by the Science \& Technology Innovation 2030 Major Program Project No. 2021ZD0150100, National Natural Science Foundation of China Project No. U21A20427, Project No. WU2022A009 from the Center of Synthetic Biology and Integrated Bioengineering of Westlake University, Project No. WU2023C019 from the Westlake University Industries of the Future Research. Finally, we thank the Westlake University HPC Center for providing part of the computational resources, and Project No. 23-407-3-29 from the Shenyang Science and Technology Program.

% \clearpage  % TODO REVIEW/FINAL: This \clearpage needs to be removed from both review and camera-ready versions.

\bibliographystyle{splncs04}
\bibliography{main}

\appendix
\newpage

\section{Pseudocode}
\label{app:pseudocode}

The pseudocode for MC-CoT, as outlined in Algorithm~\ref{alg:mc-cot}, is structured into three specialized functions, each designed to fulfill a specific aspect of the procedure. The first function, named \texttt{rationale\_generation}, is dedicated to generating rationales by considering the given question and image. Following this, the \texttt{answer\_inference} function is tasked with producing the answer, leveraging the input question, image, and the rationale generated earlier. Importantly, the third function, \texttt{voting}, is a crucial component of the framework, implementing the voting mechanism. This mechanism is essential in both the \texttt{rationale\_generation} and \texttt{answer\_inference} stages. During execution, these functions are repeated multiple times, specifically $N_r$ and $N_a$ times, respectively, to create a range of rationales and answers.

The process culminates in the selection of the final rationale and answer through a voting system. This system filters the most appropriate rationale and answer from the pool of generated options. This approach ensures that the final outputs are the most fitting and relevant responses, reflecting the collective judgment of the multiple iterations.

\begin{figure}[h]
\centering
\begin{algorithm}[H]
\caption{Pseudocode of MC-CoT}
\label{alg:mc-cot}
\definecolor{codeblue}{rgb}{0.25,0.5,0.5}
\definecolor{codekw}{rgb}{0.85, 0.18, 0.50}
\lstset{
    backgroundcolor=\color{white},
    basicstyle=\fontsize{7.5pt}{7.5pt}\ttfamily\selectfont,
    columns=fullflexible,
    breaklines=true,
    captionpos=b,
    commentstyle=\fontsize{7.5pt}{7.5pt}\color{codeblue},
    keywordstyle=\fontsize{7.5pt}{7.5pt}\color{codekw},
}
\begin{lstlisting}[language=python]
def rationale_generation(text, image, N_r):
    # generate rationale
    rationales = [f(text, image) for i in range(N_r)]
    # voting for the best rationale
    rationale = voting(rationales)
    return rationale

def answer_inference(text, image, rationale, N_a):
    # generate answer
    answers = [f([text, rationale], image) for i in range(N_a)]
    # voting for the best answer
    answer = voting(answers)
    return answer

def voting(logits, alpha=0.5):
    logits = torch.stack(logits, dim=0)
    mean_logits = torch.mean(logits, dim=0)
    std_logits = torch.std(logits, dim=0)
    weights = 1 / (1 + std_logits)
    weighted_logits = torch.sum(weights * logits, dim=0) / torch.sum(weights, dim=0)
    final_logits = alpha * mean_logits + (1 - alpha) * weighted_logits
    return final_logits
\end{lstlisting}
\end{algorithm}
\end{figure}

\section{Predicted Examples}
\label{app:examples}

We present additional predicted examples in Figure \ref{fig:predict_example}. Furthermore, we include two representative questions without image context to assess the models' language reasoning capabilities. MC-CoT consistently produces high-quality rationales and accurately answers questions.

\begin{figure*}[h!]
\centering
\includegraphics[width=0.88\textwidth]{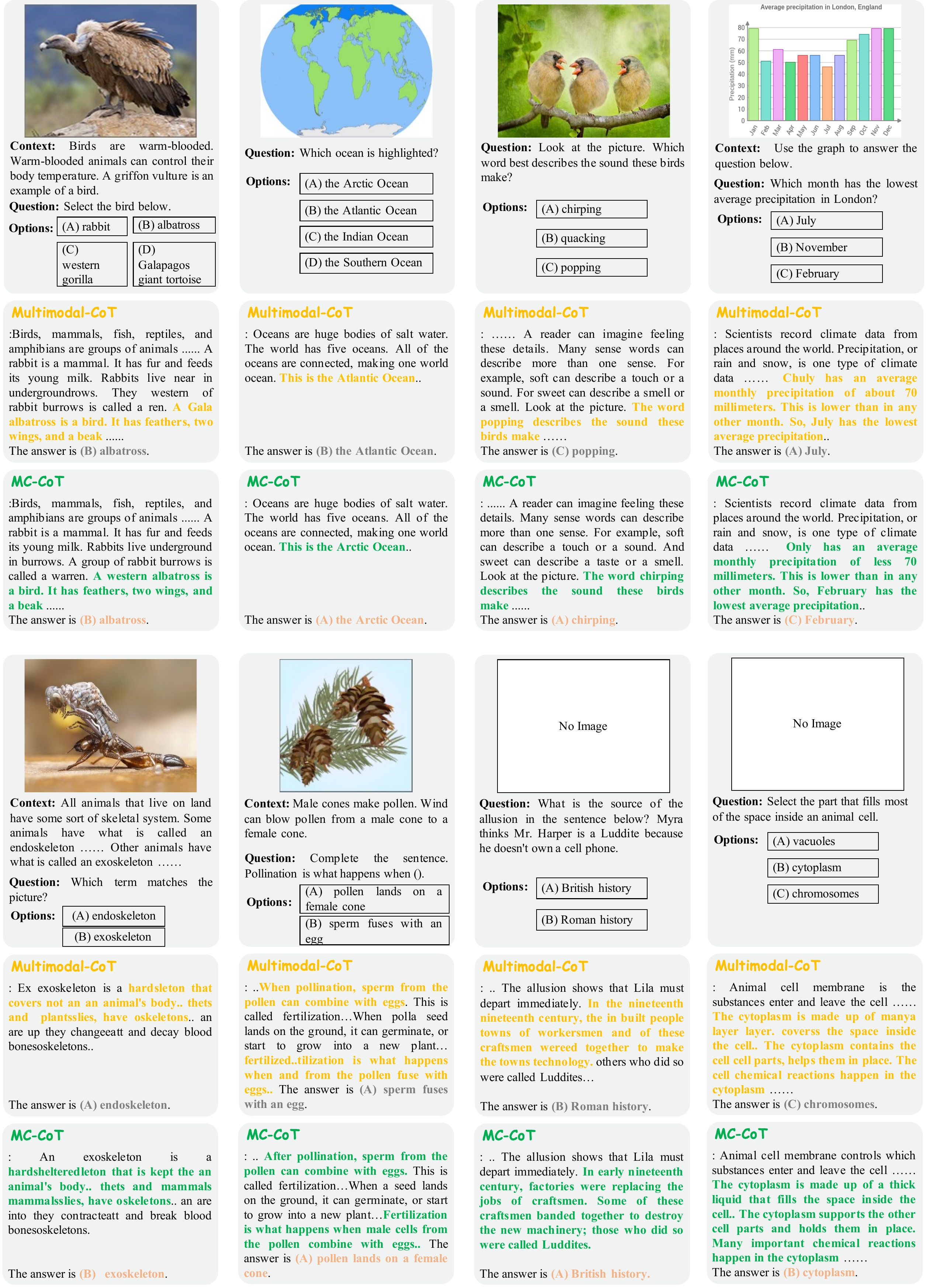}
\caption{Comparison on more predicted examples.}
\label{fig:appendix_examples}
\end{figure*}

\end{document}